\documentclass{article} 
\usepackage{iclr2020_conference,times}

\usepackage{amsmath,amsfonts,bm}

\def\eqref#1{equation~\ref{#1}}

\def\1{\bm{1}}

\DeclareMathAlphabet{\mathsfit}{\encodingdefault}{\sfdefault}{m}{sl}
\SetMathAlphabet{\mathsfit}{bold}{\encodingdefault}{\sfdefault}{bx}{n}

\usepackage{hyperref}
\usepackage{url}
\usepackage{caption} 
\usepackage{graphicx}
\usepackage{amsmath}

\title{The Dynamical Gaussian Process Latent Variable Model in the Longitudinal Scenario}

\author{Thanh Le \\
College of Information Science and Technology\\
The Pennsylvania State University\\
State College, PA 16803, USA \\
\texttt{\{txl252\}@psu.edu} \\
\And
Vasant Honavar \\
College of Information Science and Technology\\
The Pennsylvania State University\\
State College, PA 16803, USA \\
\texttt{\{vhonavar\}@ist.psu.edu} \\
}

\iclrfinalcopy 
\begin{document}

\maketitle

\begin{abstract}
The Dynamical Gaussian Process Latent Variable Models provide an elegant non-parametric framework for learning the low dimensional representations of the high-dimensional time-series. Real world observational studies, however, are often ill-conditioned: the observations can be noisy, not assuming the luxury of relatively complete and equally spaced like those in time series. Such conditions make it difficult to learn reasonable representations in the high dimensional longitudinal data set by way of Gaussian Process Latent Variable Model as well as other dimensionality reduction procedures. In this study, we approach the inference of Gaussian Process Dynamical Systems in Longitudinal scenario by augmenting the bound in the variational approximation to include systematic samples of the unseen observations. We demonstrate the usefulness of this approach on synthetic as well as the human motion capture data set. 
\end{abstract}

\section{Introduction}
\begin{figure}[ht]
    \centering
    \includegraphics[scale=0.4]{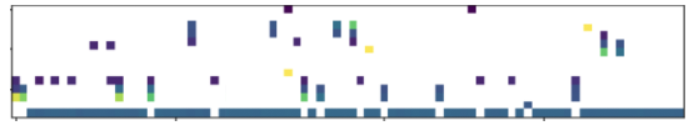}
    \caption{Example of observational density in a subset of longitudinal variable in a patient's Electronic Health Record. Each row represent a clinical variable over a period of 3 years}
    \label{fig:f1}
\end{figure}

While it isn't trivial to find an unified definition of multivariate longitudinal data; the one definition being alluded to in \cite{Pullenayegum2016LongitudinalDesign} is the type of data being discussed in this work. Longitudinal designs track a repeated set of variables in experimental subjects over periods of time; however, unlike time series, which are often characterized by regular intervals of n-dimensional observations, longitudinal setups present inconsistent sampling frequencies and only a small subset of variables may be observed at any given time. Many scientific and real-world observational data are longitudinal in nature due to various practical and experimental constraints. As a motivating example, Fig. \ref{fig:f1} visualizes a subset of temporal clinical variables in a real-world patient's Electronic Medical Record over a 3 year period. The high-dimensionality and massive amount of unascertained temporal entries in this type of data pose difficult challenges for current statistical and machine learning methods.  

Recently, \cite{Damianou2011, Damianou2014VariationalModels} developed the Dynamical Gaussian Process Latent Variable Model (D-GPLVM), based on the Bayesian Gaussian Process Latent Variable Model (Bayesian GPLVM) in \cite{Titsias2010} to learn the low-dimensional representation of a multivariate time series. This model offers an appealing solution for multivariate time series dimensionality reduction as it enables capturing of the non-linearity in the data via the use of kernels and is non-parametric (\cite{Rasmussen2006, Lawrence2005}). VGPDS is a generative model; assuming a fully observed dataset, one might generate observations at any arbitrary time using only time as inputs. Building on the work of (\cite{Girard2003,Girard2004ApproximateModels}), \cite{Damianou2016VariationalProcesses,Damianou2014VariationalModels} established the semi-described and semi-supervised blueprints for dealing with the scenario where inputs and outputs are uncertain or missing. However, these setups still require a prescription of fully observed samples. As shown in the motivating example, fully observed data are not always available. It is reasonable to believe that the ability to learn "good" representations and associated mappings for high-dimensional longitudinal data will enable the leverages of many existing techniques and insight in the rich multivariate time-series toolbox. Given this background, the motivation of this study is to enable the learning of VGPDS in the longitudinal scenario where observations are sparse both in the temporal and the feature dimension.

The remainder of this study proceeds as follows. In the next section, we provide backgrounds on the Dynamical GPLVM and the variational inference framework. We then provide an approach for learning in the longitudinal scenario by sampling unobserved variables first through a linear multitask model and more efficiently, through the Sparse Process Convolution framework. We conclude this study by demonstrating experimental results and some discussion.
%
%
\section{Background}
\subsection{Gaussian Processes}
Gaussian Process (GP), a widely used method in many machine learning applications (\cite{Rasmussen2006}), is a flexible Bayesian non-parametric model and is the building block of the Gaussian Process Latent Variable Model. In GP, we model a finite set of random function variables $f=[f(x_1),...,f(x_N)]^T$ as a joint Gaussian distribution $f \sim \mathcal{GP}(\mu, K)$ where the covariance matrix $K$ is evaluated using choices of kernel functions.

In the GP Regression, the goal is to predict the response $y^*$ of a new input $x^*$, given a training set $\{(x_i,y_i)\}^N_{i=1}$ of $N$ training samples. The response variable $y_i$ is modeled as the function value $f(x_i)$ corrupted by noise $y_i \sim \mathcal{N}(f(x_i), \sigma^2)$. Given the joint probability of the response variables and the latent function $p(y,f) = p(y|f)p(f)$, the distribution of the latent function value $f^*$ is a Gaussian distribution with mean and variance
\begin{equation}
    \begin{split}
        \mu(x^*) &= k_{x^*X}(\sigma^2I+K_{XX})^{-1}y \\
        var(x^*) &= k_{x^*x^*} - k_{x^*X}(\sigma^2I + K_{XX})^{-1}k_{Xx^*} \\
    \end{split}
\end{equation}
where $k_{x^*X} = k(x^*, X)$ is the covariance between the new input $x^*$ and the $N$ training sample evaluated by the kernel function $k$.
\subsection{Gaussian Process Latent Variable Model (GPLVM)}
GPLVM was first conceived as an approach to facilitate visualization by mean of dimensionality reduction \cite{Lawrence2004} and can be seen as a non-linear extension of the Probabilistic PCA. The major difference between GPLVM and its standard GP regression counter-part is whether the input variable $X$ is given at training time. The goal of GPLVM is to learn the low dimensional representation $X^{N\times Q}$ of the data matrix $Y^{N\times D}$ where $Q \ll D$. The mapping $f:X\rightarrow Y$ is a nonlinear function with Gaussian Process (GP) prior $f \sim \mathcal{GP}(0,K)$. The generation process of the $i^{th}$ training sample $y_i$ is therefore

\begin{equation}
    y_i = f(x_i) + \epsilon
\end{equation}

GPLVM allows the flexibility for specifying prior over the latent space $X$; one might utilize GPLVM without specifying any prior assumption, however this lack of prior is equivalent to maximizing the log marginal likelihood that is prone to over fitting (\cite{Li2016}).  Up until \cite{Titsias2010}, the standard approach in learning GPLVM is to find the MAP estimate of $X$ (\cite{Lawrence2005}) whilst jointly maximizing with respect to the hyper-parameters. Over the years, there has been various efforts to study GPLVM in different learning scenarios and corresponding methodologies enabling GPLVM to model different systems. Of our particular interest is GPLVM with dynamical prior to model multivariate time series data (\cite{Damianou2011, Wang2006}).

\subsubsection{Bayesian-GPLVM}
\cite{Titsias2009VariationalProcesses,Titsias2010} provided a full Bayesian treatment of the GPLVM - a GP prior based on auxiliary inducing points was introduced so that the variational Bayes approach was tractable. The latent variables were then variationally integrated out and a close-form lower bound on the marginal likelihood computed. The original purpose of inducing points in \cite{Csato2001} was to speed up computation. The marginal likelihood of the data $p(Y) = \int p(Y|X)p(X)dX$ is intractable because $X$ appears nonlinear inside the covariance matrix $K_{NN} + \beta^{-1}I_N$. A variational distribution $q(X)$ is introduced to approximate the true posterior distribution $p(X|Y)$. The chosen variational distribution in the i.i.d case have a factorized Gaussian form

\begin{equation}
q(X)=\prod^N\mathcal{N}(x_{n}|\mu_{n}, S_{n})
\end{equation}

resulting in the Jensen's lower bound on the $log\ p(Y)$ taking the form

\begin{equation}
\begin{split}
F(q)  = \tilde{F}(q) - KL(q(X)|p(X))
\end{split}
\end{equation}
the negative KL divergence between the variational posterior distribution q(X) and the prior distribution $p(X)$ and can be computed analytically while the first term breaks down to separate computations at each $d^{th}$ dimension.
\begin{equation}
\begin{split}
    \tilde{F}(q) &= q(X)log p(Y|X)dX \\
    &=\sum_{d=1}^D \int q(X) log p(y_d|X)dX=\tilde{F}_d(q)
\end{split}
\label{eq:boundbreakdown}
\end{equation}
The intractable integration $log p(y_d|X)$ involves in $\tilde{F}_d(q)$ can then be approximated using inducing points. For each vector $f_d$, a set of M inducing variables $u_d$ is introduced; $u_d's$ are evaluated at a set of inducing locations given by $Z \in R^{M\times Q}$.  $U$ are simply the function points drawn from the same conditional prior, augmenting the joint probability model in (6) 
\begin{equation}
p(y_d, f_d, u_d|X,Z) = p(y_d|f_d)p(f_d|u_d,X,Z)p(u_d|Z)
\end{equation}
The likelihood $p(y_d|X) $ can be computed from the augmented model by marginalizing out $(f_d,u_d)$ for any value of the inducing inputs $Z$. This allows $p(f_d|X)$ to be computed by $ q(f_d,u_d)=p(f_d|u_d,X)\phi (u_d) = p(f_d|u_d)\phi (u_d)$, which is tractable. The bound for the data can then be fully specified by the Psi statistics $\Psi_0=Tr(\big<K_{NN}\big>_{q(X)})$, $\Psi_1=\big<K_{NM}\big>_{q(X)}$, $\Psi_2=\big<K_{MN}K_{NM}\big>_{q(X)}$ where $\big<\cdot \big>_{q(X)}$ denotes the expectation under the variational distribution $q(X)$ \cite{Lawrence2007}.  Note that the above statistics involve convolution of the covariance function with a Gaussian density and can only be analytically obtained for some standard kernels. An attractive integral part of this is the ability to automatically determine the latent dimensionality of a given dataset by Automatic Relevance Determination (ARD).

\subsubsection{Variational Gaussian Process Dynamical Systems}
A dynamical prior can be imposed over the $X$ in the GPLVM to enable modeling of dynamical system \cite{Lawrence2007, Damianou2011}. In the multivariate time series data $\{y_n, t_n\}_{n=1}^N$, where $y_n \in \rm I\!R^D$ is a d-dimensional observation at time $t_n \in \rm I\!R^+$. The system could be summarized as follow

\begin{equation}
\begin{split}
    x_q(t) \sim \mathcal{GP}(0, k_x(t_i, t_j)), q = 1,...,Q  \\
    f_d(x) \sim \mathcal{GP}(0, k_f(x_i, x_j)), d = 1,...,D
\end{split}
\end{equation}
The kernel functions $k_x$ and $k_f$ are parameterized by $\theta_x$ and $\theta_f$ respectively. The choice of $k_x$ to be indefinitely differentiable function such as the square exponential (RBF) allow generation of a smooth path in the latent space. Indeed, the major difference between the Bayesian GPLVM in \cite{Titsias2010} and Dynamical GPLVM in \cite{Damianou2012} is the use of dynamical prior. As a result, the derivations of the lower bound are similar with the exception of the KL divergence being $KL(q(X)||p(X|t))$ and $X$ are coupled temporally leading to the factorization on $q(X)$:
\begin{equation}
    q(X) = \prod^Q \mathcal{N}(x_q|\mu_q, S_q)
    \label{eq:qXfullS}
\end{equation}
This results in a full-rank covariance matrix $S_q$ with $N^2$ parameters.  Following the re-parameterization trick in \cite{Opper2009TheRevisited}, however, reduces the number of parameters to that of the standard Bayesian Gaussian Process. Specifically $\mu_q$ and $S_q$ can be parameterized by the   $\mu_q=K_t\bar{\mu}_q$ and $S_q = (K_t^{-1} + \Lambda_q)^{-1}$ where $\bar{\mu}_q$ and $\Lambda_q$  consist of $Q \times N$ free parameters. 
%
%
\section{Gaussian Process Latent Variable Model in the Longitudinal Scenario}
In this section, we will first show a simple sampling procedure of the unseen observations would resolve the difficulty in learning the dynamical representation in the longitudinal scenario. We will then introduce a more computationally efficient alternative in the following subsection.

In the ideal situation, the dynamical structure is properly propagated from $X$ to $Y$; nevertheless, there are no fully observed data and the variational approximation described in the GPLVM is, therefore, non-trivial. In the case where a global dynamical covariance is optimal, such covariance structure is also optimal for individual data dimensions. This is true by construction, the computation of the lower bound of the marginal likelihood breaks down to separate computations over each data dimension in (\ref{eq:boundbreakdown}). Furthermore, the quantity of the lower bound, after being optimally eliminated, boils down to  computing the $\Psi$ statistics which are expectations of different covariances under the variational distribution, $q(X)$ in (\ref{eq:qXfullS}) where both $\mu_q$ and $S_q$ involve parameterization by the dynamical kernel $K_t$.  
This supports the direct introduction of the dynamical covariance to estimate the sampling distribution for the unseen observations $y_d^*$ in the $d^{th}$ dimension at time $t^*$  
\begin{equation}
\label{eq:mosamples}
y_d^* \sim \mathcal{N}(\mu_{mo}^*, \Sigma^*_{mo})
\end{equation}
where $\mu_{mo}^*, \Sigma^*_{mo}$  are the mean and covariance of a multitask Gaussian Process estimated over the observed outputs with the shared covariance function, $K_t$, in the heterotopic setting (i.e., the training data do not align). Here, each data dimension is treated as an individual task in a multi-task problem. As we will discuss, we define a new objective on the composite log-likelihood over the observed data
\begin{equation}
\label{eq:compositebound}
F^*(q) \geq \mathcal{L}_{mo}(y^o; K_t) + \tilde{\mathcal{F}_*}(q) - KL(q||p)
\end{equation}
The multitask model log-likelihood $\mathcal{L}_{mo}$ is computed over the observed data using the global dynamical covariance $K_t$; and $\tilde{\mathcal{F}}^*(q)$ is the dynamical model log-likelihood computed over the fully sampled data $\tilde{\mathcal{F}}^*(q) = \sum_{d=1}^D \tilde{\mathcal{F}}_d^* = \sum_{d=1}^D q(X) log\ p(y^*_d, y^o_d|X)dX$. The choice of juxtaposing two multi-task models and optimizing them simultaneously were informed by the different purpose of each model. The multi-output regression model is used as an adaptive imputation of the unobserved $Y^*$ at every optimization step using the current best approximation of the global dynamical covariance. At the same time, the Dynamical GPLVM attempts to find an optimal dynamical covariance as well as the cross-covariance structures using the sampled and observed data. As a result, the inputs to the multi-output regression and the Dynamical GPLVM are time, and the hidden variable $X$, respectively. 

\subsubsection{Simple sampling using multi-output Gaussian Process}
One of the most straightforward choices to establish the sampling distribution is to use the Linear Model of Coregionalization (LMC) \cite{Bonilla2008Multi-taskPrediction,Li2016}:
\begin{equation}
\begin{split}
	\mu_{mo}^* &=(K_f \otimes K_t(t^*,t^o))^T(\Sigma^*_{mo})^{-1}y^o\\
    \Sigma^*_{mo} &=K_f\otimes K_t + D \otimes I
\end{split}
\label{eq:LMC}
\end{equation}
in which, $K_f=\Phi\Phi^T$ is the task covariance matrix. In the case where data dimensions are known a priori to be independent, $K_f$ can be fixed to be the identity matrix $K_f = I_D$. Regardless of data dimension dependencies, initializing $K_f=I_D$ allows the unseen data to first be sampled independently. 

\begin{figure}
    \centering
    \includegraphics[scale=0.55]{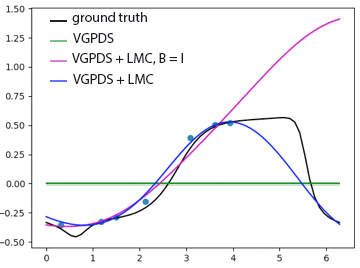}
    \caption{Visualization of predictive mean generated using only time as input on a sample dimension of the synthetic data where observation density is 0.1 . The blue dots represent observations available in this data dimension. The auxiliary LMC model enables VGPDS to explicitly capture the dependency among tasks in sparse dataset}
    \label{fig:synsample}
\end{figure}

Fig. \ref{fig:synsample} demonstrates the ability of this approach to learn a representation and associated mapping to produce reasonable predictive means in a synthetic longitudinal dataset. The pitfall of this simple approach is the complexity during training being dominated by the LMC with the naive implementation of typical cubic complexity $O(N^3D^3)$ or a reduced complexity of  $O(NDM^2P^2)$ with a couple approximations such that $M\ll N$ and $P < D$  (For more details on the approximations and complete derivation of $\mathcal{L}_{mo}$ over the observed data, refer to \cite{Bonilla2008Multi-taskPrediction}).
\subsubsection{More efficient sampling via sparse convolved Gaussian Processes}
In this subsection, we will establish a more efficient construction of the sampling distribution in \ref{eq:mosamples} using the Sparse Process Convolution framework. We will first re-emphasize the Dynamical GPLVM is a multitask Gaussian Process where the temporally correlated input $X$ is hidden \cite{Titsias2009VariationalProcesses}. Though, unlike the LMC whose multitask covariance is captured by the Kronecker product between the coregionalization matrix, $K_f$, and the input covariances, the Bayesian GPLVM follows the Process Convolution formalism to model the dependencies among the individual tasks \cite{Alvarez2011ComputationallyProcesses,Alvarez2009VariationalProcesses,Alvarez2011EfficientKernels}.  
In the Process Convolution framework, each function $f_d(x)$, the noiseless version of $y_d$, can be expressed through a convolution integral between a smoothing kernel $G_d$ and the shared latent function $u$ 
\begin{equation}
f_d(x)=\int_{\mathcal{X}}G_d(x-z)u(z)dz
\label{eq:fdconv}
\end{equation}
While it is possible to have multiple latent functions $u$'s each with their own set of smoothing kernel; for simplicity, we will assume the one latent function as shown in (\ref{eq:fdconv}). Under the same independence assumptions in LMC, if $u$ is chosen to be independent white Gaussian noise process with a general covariance, $k(z',z)$, then the (cross-)covariances can be computed
\begin{equation}
\begin{split}
	cov[f_d(x), f_d'(x')] &= \int_{\mathcal{X}} G_d(x-z) G_d'(x'-z)k(z,z')dz'dz \\
	cov[f_d(x), u(z)] &= \int_{\mathcal{X}} G_d'(x'-z)k(z,z')dz'
\end{split}
\end{equation}
If the smoothing kernel, $G_d$, is taken to be the Dirac delta function then resulting model turns out to be the LMC. \cite{Alvarez2009VariationalProcesses,Alvarez2011ComputationallyProcesses} showed that instead of drawing a sample from $u(z)$, they could summarize $u(z)$ by drawing samples from its finite representation, i.e., $u=[u(z_1),...,u(z_M)]^T$ .... . Each function $f_d$ in (\ref{eq:fdconv}) can then be reasonably approximated by
\begin{equation}
f_d(x) \approx \int_{\mathcal{X}} G_d(x-z)E[u(z)|u]dz
\end{equation}
In this study, we will assume $u$ to already be reasonably smooth and $f_d$s are independent conditional on $u$. The likelihood of $f$ is
\begin{equation}
\begin{split}
p(f|u,Z,X,\theta) &= \mathcal{N}(f| K_{f,u}K_{u,u}^{-1}u, K_{f,f}-K_{f,u}K_{u,u}^{-1}K_{u,f}) \\
&= \prod^D \mathcal{N}(f| K_{f_d,u}K_{u,u}^{-1}u, K_{f_d,f_d}-K_{f_d,u}K_{u,u}^{-1}K_{u,f_d}) \\
\end{split}
\end{equation}
The full multitask covariance matrix is now replaced by its low rank approximation  $K_{f,u}K_{u,u}^{-1}K_{u,f}$ in all entries except in the diagonal block corresponding to $K_{f_d,f_d}$,  the overall computational complexity is improved. We can approach constructing the sampling distribution in (\ref{eq:mosamples}) using this method. Follows the same arguments in the previous subsection, we can establish the sampling distribution as follow: 
\begin{equation}
\begin{split}
	\mu_{mo}^* &= K_{f_*,u}A^{-1}K_{u,f_*}(D+\Sigma^*_{mo})^{-1}y_o\\
    \Sigma^*_{mo} &=K_{f_*, f_*} - K_{f_*,u}K_{u,u}^{-1}K_{u,f_*} + K_{f_*,u}A^{-1}K_{u,f_*} + \Sigma_*
\end{split}
\end{equation}
$A=K_{u,u} + K_{u,f}(D+\Sigma)^{-1}K_{f,u}$, the relevant multitask likelihood over the observed output sharing the same covariance parameters as the Dynamical model: 
\begin{equation}
\mathcal{L}_{mo} \propto  -\frac{1}{2} log |K_{u,u}| - \frac{1}{2} log |A| - \frac{1}{2} tr\big[ D^{-1}yy^T\big] + \frac{1}{2}tr\big[ D^{-1}K_{f,u}A^{-1}D^{-1}yy^T\big]
\end{equation}
Like before, this multitask Gaussian Process Model share a set of parameters from the Dynamical model at every step of the optimization. We will assume $K_{u,u}$ is the shared covariance functions. 

 The computational complexity of this approach is on the same order of the original Dynamical GPLVM.

%
%
\section{Experiments}
In this section, we consider simulated and real-world data to demonstrate the ability of the described approach to learning reasonable latent representation as well as to impute the unobserved data in the longitudinal setting. The synthetic data were created by feeding time as input into generating processes unknown to the competing dynamical models, the final output is corrupted by Gaussian noise. For the real-world data, a subset of the same motion capture dataset corresponds to the walking motion of a human body represented as a set of 59 joint locations. During each trial for the synthetic and MOCAP dataset (\cite{CMU2001MOCAPDatabase}), an increasing amount of data were masked as missing; equivalently, the observation matrix density was decreasing. The models were evaluated based on reconstruction error using only time as input. In these experiments, 5 methods are considered - The Dynamical GPLVM (\cite{Damianou2011}), deepGP with dynamical prior (\cite{Damianou2012}),  Variational GP Longitudinal Model using single and multi-output data augmentation (SO-VGPLS, MO-VGPLS respectively), and Nearest Neighbor (NN). Even though NN is not a temporal model, the reconstruction error was shown to compare this popular imputation approach to the others in the result.

\subsubsection*{Synthetic dataset}
In this experiment, RBF kernel was chosen to be the dynamical kernel. The reconstruction error in Table.1 showed the performance of the standard Dynamical GPLVM quickly degraded as the observed data became progressively thinner; at around 30\% observation density, the D-GPLVM appeared to no longer infer informative structure from the synthetic dataset. On the contrary, VGPLS were able to reproduce the generated data most accurately among all the examined methods even at extremely low observation density such as that found in the motivating example.

\begin{table*}[ht]
\centering
\resizebox{\textwidth}{!}{%
\begin{tabular}{l|lllllllll}
Obs Density       & 0.9                & 0.7                & 0.5                & 0.4                & 0.3                & 0.2                & 0.1                 \\ 
\hline
NN                & 22.4(0.9)          & 29.9(4.6)          & 62.0(6.0)          & 65.2(5.6)          & 80.1(6.4)          & 107(1.1)           & 123(4.5)            \\
\textbf{D-GPLVM}    & \textbf{21.5(1.6)} & \textbf{48.2(1.4)} & \textbf{75.4(1.5)} & \textbf{87.2(2.2)} & \textbf{128(0.4)}  & \textbf{128(0.4)}  & \textbf{129(0.1)}   \\
deepGP            & 17.6(1.1)          & 33.0(2.3)          & 57.4(1.3)          & 67.4(0.9)          & 85.9(0.8)          & 104(3.8)           & 128(1.0)            \\
\textbf{VGPLS} & \textbf{14.2(0.7)} & \textbf{19.4(2.8)} & \textbf{28.8(1.2)} & \textbf{32.4(2.5)} & \textbf{36.1(0.7)} & \textbf{53.4(1.9)} & \textbf{59.0(7.6)} 
\end{tabular}
}
\caption{Reconstruction Absolute Sum of Error using only time as input. The experiment was repeated 3 times and the mean (stdev.) values were recorded}
\end{table*}
\subsubsection*{Human motion capture data}
In this experiment, we were interested in not only the error rate of reconstruction but also the quality of the latent representation learned using the sparse dataset. The dynamical kernel chosen to model the walking movement was to be the periodic kernel as this type of motion capture consists of repeated joint movements.

Table 2 demonstrated the same performance advantage of VGPLS over the standard D-GPLVM shown in the MOCAP data
\begin{table}[h]
\centering
\begin{tabular}{l|llll}
Observation Density & 1.0           & 0.8           & 0.6          & 0.4           \\ 
\hline
D-GPLVM (RBF)    & 84.2          & 252           & 332          & 332           \\
deepGP (RBF)   & \textbf{82.3} & 143           & 201          & 320           \\
VGPLS (RBF) & 83.1          & \textbf{84.9} & \textbf{135} & \textbf{221} 
\end{tabular}
\caption{Reconstruction Absolute Sum of Error Using only time as input in temporal models}
\end{table}

%
%
\section{Discussions}
In this study, we provide a simple approach for learning the Dynamical Gaussian Process Latent Variable Model in the longitudinal scenario. This approach proposed the sharing of the parameters between the Dynamical Model and another Multitask Gaussian Process model to enable sampling of unseen observations at every step of the gradient-based optimization of the variational lower bound.  Furthermore, by leveraging the Sparse Process Convolution framework, this approach learns the latent representation in longitudinal setting with minimal computation overhead. The learned representation of the longitudinal data arguably exhibits properties that is more desirable in comparison to the original sparse dataset, namely, a complete and low dimensional with fully specified covariance structures. The feasibility of the method was demonstrated in synthetic and human motion capture data.

We chose to use longitudinal scenario instead of explicitly longitudinal data as this approach could be apply to different Gaussian Process Latent Variable Models in which a prior is well specified over the latent space and a large portion the data is unobserved. Finally, in regard to our motivating example, through experimentation, we did not find the proposed model adequate or suitable for addressing the needs in this type of dataset due to the limitation of the temporal assumption. It is difficult, and perhaps unreasonable to impose a suitable dynamical structure over health observations which are very far apart. Nevertheless, this model could be useful for other use case where the kernel dynamical assumption suffices.

\bibliography{references}
\bibliographystyle{iclr2020_conference}

\end{document}